\documentclass[review,12pt,3p]{elsarticle}
\usepackage{graphicx}

\usepackage{subcaption}

\usepackage{color}

\begin{document}

\title{Generic Feature Learning for Wireless Capsule Endoscopy Analysis}

\author[label1,label2]{Santi Segu\'i\corref{cor1}}
\address[label1]{Dept. Matem\`atica Aplicada i An\`alisi, Universitat de Barcelona, Barcelona}
\address[label2]{Computer Vision Center (CVC), Barcelona, Spain\fnref{label4}}
\cortext[cor1]{Corresponding author Telf: +34 93 402 16 31}

\ead{santi.segui@ub.edu}

\author[label5]{Michal~Drozdzal}
\address[label5]{Medtronic GI, Yoqneam, Israel}

\author[label1]{Guillem Pascual}
\author[label1,label2]{Petia Radeva}
\author[label3]{Carolina Malagelada}
\address[label3]{Digestive System Research Unit, Hospital Vall d'Hebron, Barcelona, Spain}
\author[label3]{Fernando Azpiroz}
\author[label1,label2]{Jordi Vitri\`a}

\maketitle

\begin{abstract}
The interpretation and analysis of the wireless capsule endoscopy recording is a complex task which requires sophisticated computer aided decision (CAD) systems in order to help physicians with the video screening and, finally, with the diagnosis. Most of the CAD systems in the capsule endoscopy share a common system design, but use very different image and video representations. As a result, each time a new clinical application of WCE appears, new CAD system has to be designed from scratch. This characteristic makes the design of new CAD systems a very time consuming. Therefore, in this paper we introduce a system for small intestine motility characterization, based on Deep Convolutional Neural Networks, which avoids the laborious step of designing specific features for individual motility events. Experimental results show the superiority of the learned features over alternative classifiers constructed by using state of the art hand-crafted features. In particular, it reaches a mean classification accuracy of 96\% for six intestinal motility events, outperforming the other classifiers by a large margin (a 14\% relative performance increase).

\end{abstract}


\section{Intoduction}
\label{secIntro}

Wireless Capsule Endoscopy (WCE) is a tool designed to allow for inner-visualization of entire gastrointestinal tract. It was developed in the mid-1990s and published in 2000 \cite{iddan2000wireless}. 
The invention is based on a swallowable capsule, equipped with a light source, camera, lens, radio transmitter and battery, that is propelled by the peristalsis along all GastroIntestinal (GI) tract, allowing the full visualization of it from inside without pain and sedation.

This technology was positively accepted by the gastroenterology community, because it allows minimally invasive inspections, even of those parts of the intestine that are not accessible by classical means. 
Nowadays, despite the inability to control the motion and position of the capsule in the GI tract, this device is considered the gold standard technique for the diagnosis of  bleeding \cite{eliakim2004wireless,mustafa2013small}, and it is becoming very popular investigation tool for  possible future diagnosis of particular diseases such as tumors \cite{tumor2006,tumor2010}, chronic abdominal pain \cite{YangPlosOne} and motility disorders \cite{malagelada2012functional}.

\begin{figure}
        \centering
        \begin{subfigure}[b]{0.08\textwidth}
                \includegraphics[width=\textwidth]{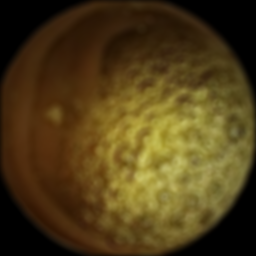}
                \caption{}
                \label{fig:gull}
        \end{subfigure}%
        ~ 
        \begin{subfigure}[b]{0.08\textwidth}
                \includegraphics[width=\textwidth]{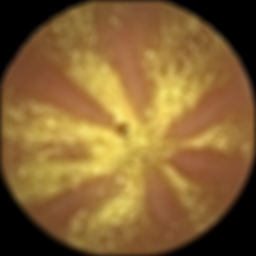}
                \caption{}
        \end{subfigure}
        ~
        \begin{subfigure}[b]{0.08\textwidth}
                \includegraphics[width=\textwidth]{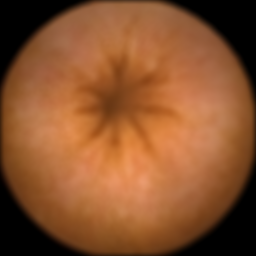}
                \caption{}
        \end{subfigure}
        ~
        \begin{subfigure}[b]{0.08\textwidth}
                \includegraphics[width=\textwidth]{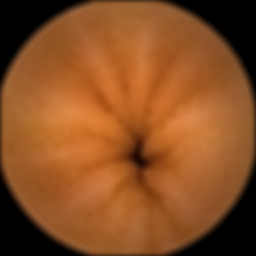}
                \caption{}
        \end{subfigure}
        ~
        \begin{subfigure}[b]{0.08\textwidth}
                \includegraphics[width=\textwidth]{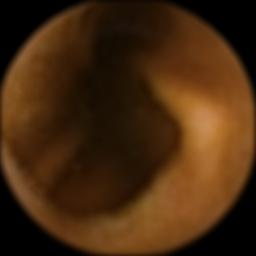}
                \caption{}
        \end{subfigure}
        ~
        \begin{subfigure}[b]{0.08\textwidth}
                \includegraphics[width=\textwidth]{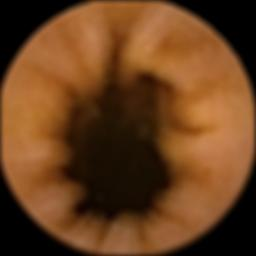}
                \caption{}
        \end{subfigure}
        
        \begin{subfigure}[b]{0.08\textwidth}
                \includegraphics[width=\textwidth]{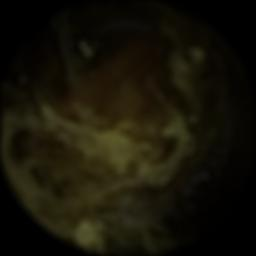}
                \caption{}
        \end{subfigure}%
        ~ 
        \begin{subfigure}[b]{0.08\textwidth}
                \includegraphics[width=\textwidth]{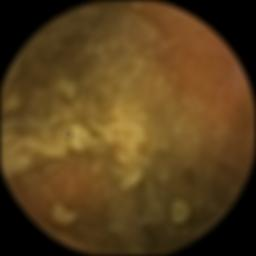}
                \caption{}
        \end{subfigure}
        ~
        \begin{subfigure}[b]{0.08\textwidth}
                \includegraphics[width=\textwidth]{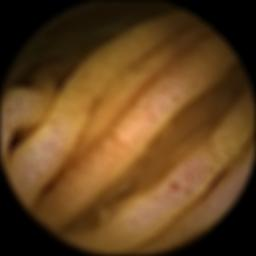}
                \caption{}
        \end{subfigure}
        ~
        \begin{subfigure}[b]{0.08\textwidth}
                \includegraphics[width=\textwidth]{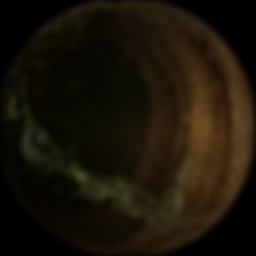}
                \caption{}
        \end{subfigure}
        ~
        \begin{subfigure}[b]{0.08\textwidth}
                \includegraphics[width=\textwidth]{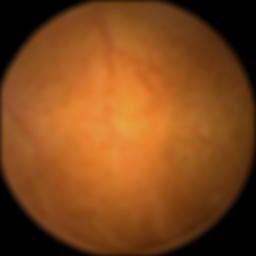}
                \caption{}
        \end{subfigure}
        ~
        \begin{subfigure}[b]{0.08\textwidth}
                \includegraphics[width=\textwidth]{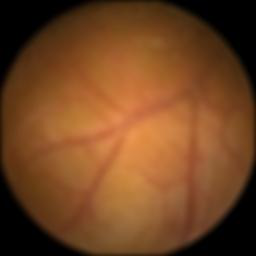}
                \caption{}
        \end{subfigure}
        
        \begin{subfigure}[b]{0.08\textwidth}
                \includegraphics[width=\textwidth]{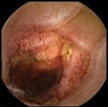}
                \caption{}
        \end{subfigure}%
        ~ 
        \begin{subfigure}[b]{0.08\textwidth}
                \includegraphics[width=\textwidth]{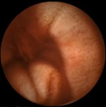}
                \caption{}
        \end{subfigure}
        ~
        \begin{subfigure}[b]{0.08\textwidth}
                \includegraphics[width=\textwidth]{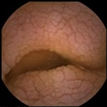}
                \caption{}
        \end{subfigure}
        ~
        \begin{subfigure}[b]{0.08\textwidth}
                \includegraphics[width=\textwidth]{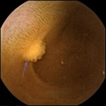}
                \caption{}
        \end{subfigure}
        ~
        \begin{subfigure}[b]{0.08\textwidth}
                \includegraphics[width=\textwidth]{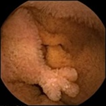}
                \caption{}
                \label{fig:mouse}
        \end{subfigure}
        ~
        \begin{subfigure}[b]{0.08\textwidth}
                \includegraphics[width=\textwidth]{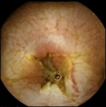}
                \caption{}
        \end{subfigure}
        
        \caption{WCE image sample: (a,b) Bubble images. (b,c,d)  Wrinkle images. (e,g) Clear blob images. (g,h) Turbid images. (i,j) Dilated images. (k,l) Wall images. (m,n) Bleeding images. (o) Celiac images. (p,q) Polyp images. (r) Crohn's image.}\label{fig:WCEsamples}
\end{figure}

The main drawbacks of WCE-based diagnosis are the length of its videos and the complexity of the images. A normal small intestine video can represent up to 8 hours of recording. This means that a single video can contain up to $57,600$ images, if 2 frames-per-second capsule is used (this number can become even larger with higher frame-rate capsules). All these frames are presented with highly variable camera orientation and perspective, since the device moves freely inside the GI tract. Moreover, sometimes the scene can be fully, or partially, hindered by the intestinal content such as bilis or food in digestion. Figure~\ref{fig:WCEsamples} allows  to fully appreciate the complexity of the endoluminal scene.  
This scene complexity together with the length of a single WCE study make the proper video analysis, by the physicians, a hard and tedious task. A recent study presented in the American Journal of Gastroneterology \cite{zheng2012detection} showed that the performance achieved by the visual analysis of physicians is far from perfect.
The detection rates of angioectasias, ulcers/erosions, polyps, and blood were $69\%$, $38\%$, $46\%$, and $17\%$, respectively. Besides, the time needed for the analysis can range from 45 minutes to several hours, depending on the expertise of the physician.

In order to  help physicians with the diagnosis, several  WCE-based CAD systems have been presented in the last years. Extensive reviews of these CAD systems can be found  in \cite{liedlgruber2011computer,belle2013biomedical, iakovidis2015software}. Generally, these systems are designed either for efficient video visualization  \cite{mackiewicz2008wireless,chu2010epitomized,iakovidis2013efficient,drozdzal2013adaptable} or to automatically detect different intestinal abnormalities such as bleeding \cite{chen2009,6497444}, polyp \cite{polypsMamonov,yuanpolyp}, tumor \cite{tumor2006}, ulcer detection \cite{ciaccio2013implementation}, motility disorders \cite{Malagelada2015,sseguiWrinkles} and other general pathologies \cite{ciaccio2010distinguishing,6051474,malagelada2012functional,Chen2013}.  For the detailed description of the WCE-based CAD systems, please refer to Section II of this paper.

Reviewing the state-of-the-art in WCE for intestinal event or pathology detectors, one observation can be made: each method uses a different, specially hand-crafted image representation (e.g. intestinal content or bleeding classifiers are based on color analysis, wrinkle-like frames detectors use  structure analysis and polyps characterization is based on structure and texture analysis). This hand-crafting process implies that each time a new event of physiological interest appears, the method for quantification of these phenomena has to be designed from scratch by an expert and therefore, the process is very time consuming.

Lately, a technology for generic feature learning in computer vision tasks, called Convolutional Neural Network (CNN), appeared and has been shown to obtain very good results in image classification tasks \cite{krizhevsky2012imagenet, DBLP:journals/corr/SzegedyLJSRAEVR14, DBLP:journals/corr/SimonyanZ14a}.  The CNN uses a hierarchy of computational layers in order to learn a mapping from an input (image) to an output (class). The main advantage of CNN is the fact that it permits to classify a variety of classes using a single model without the need for feature hand-crafting. However, in order to learn a good model, a relatively large number of labeled images is needed.

In this paper, we present a CAD system with a generic feature-learning approach for WCE endoscopy that merges CNN technology with WCE data. In order to be able to train the network, we build a large data set of 120k labeled WCE images (for details, see Section III). Our hypothesis is that a single, well trained deep network would be generic enough to classify a variety of events present in WCE recording. Therefore, the contributions of the paper can be summarized in the following points:
\begin{itemize}
  \item We adapt the architecture of CNN to the problem of endoluminal image classification. Our architecture uses as an input not only color information (RGB channels), but also priors in a form of higher-order image information (Section IV).
  \item We show that this generic system outperforms well-known computer vision descriptors (e. g. SIFT, GIST, color histogram) and establishes new state-of-the-art in intestinal event detection (Section V).
  \item Finally, we analyse and illustrate the impact of the data set size on the system performance (Section V).
\end{itemize}

\section{Automatic detection of GI events in WCE}

From computational point of view, most of the recent developments in WCE have been devoted to the design of algorithms to automatically detect different GI events or pathologies. Table \ref{tab:lit_review} presents a review of applications of WCE such as bleeding \cite{figueiredo2013computer,Sainju2014,6497444,yeh2014bleeding}, informative frames detection (visibility) \cite{sseguiTurbid,Sun2013,maghsoudi2014informative, Suenaga2014Bubbles, zhao2015general}, motility \cite{sseguiWrinkles, Malagelada2015, Drozdzal2015}, ulcers \cite{Eid2013,yeh2014bleeding}, polyps \cite{polypsMamonov,yuanpolyp,silivaPolyp2014, zhao2015general} celiac disease \cite{ciaccio2013implementation} and Crohn's disease \cite{6051474} during the last 3 years.

Most of these CAD systems are built with the common three-stage design: 1) database creation, 2) image representation that is usually defined manually, and 3) classifier. The goal of the CAD system is to predict the presence or absence of certain GI event in new WCE images. In the first step of the system design process, researchers need to build a representative database of labeled images of the problem they want to solve. From the machine learning perspective, the size of the database depends on the problem complexity. However, in medical imaging the databases are typically limited by the high costs of image labeling and data acquisition. The next step of the architecture is the design of the WCE image representation. This step relies on researcher's expertise and intuition in the data and computer vision tools. Finally, the last step of the CAD system is usually the decision making by means of a classifier. 

In the remaining part of this section, we review a variety of image representations defined for WCE data, including color, shape and texture features.

\begin{table*}[!ht]
\centering
\resizebox{12cm}{!} {
    \begin{tabular}{l c c c c c c c c c c }
    \hline
     Paper & Year &  \multicolumn{6}{c}{ Applications} & & Feature\\
      & &  Bleeding & Visibility & Motility & Ulcer & Polyps & Celiac & Crohn's\\
    \hline
    Eid et al. \cite{Eid2013} &2013 &  & &  & $\surd$   & & & & Texture\\
    Figuereido et al.\cite{figueiredo2013computer} &2013 &  $\surd$ & &  & &   & & & Color\\
    Sanju et al.\cite{Sainju2014} &2014 &  $\surd$ & &  & &   & & & Color\\
    Fu et al.\cite{6497444}                       &2014 &  $\surd$ &  & & & & & & Color\\
    Yeh et al.\cite{yeh2014bleeding}              &2014 &  $\surd$ &  & & $\surd$ & & & & Color\\
    Segui et al.\cite{sseguiTurbid}        &2012 &  & $\surd$ & & &  & & & Color and Texture\\
    Sun et al.\cite{Sun2013}        &2013 &  & $\surd$ & & &  & & & Color and Texture\\
    Maghsoudi et al. \cite{maghsoudi2014informative}  &2014 &  & $\surd$ & & &  & & & Color, Shape and Texture\\
    Suenaga et al. \cite{Suenaga2014Bubbles}  &2014 &  & $\surd$ & & &  &  & & Shape\\
    Segui et al.\cite{sseguiWrinkles}      &2014 &  &  & $\surd$ & &  & & & Shape \\
    Malagelada et al.\cite{Malagelada2015}      &2015 &  &  & $\surd$ & &  & & & Color, Shape and Texture \\
    Drozdzal et al.\cite{Drozdzal2015}      &2015 &  &  & $\surd$ & &  & & & Color, and Shape\\
    Mamonov et al.\cite{polypsMamonov}      &2013 &  & &  & & $\surd$ & & & Shape and Texture \\
    Yuan et al.\cite{yuanpolyp}             &2013 &  & &   & & $\surd$ & &  &Color, Shape and Texture \\
    Silvia et al. \cite{silivaPolyp2014}    &2014 &  & &  & & $\surd$ & & &Shape and Texture \\
    Nawarathna et al. \cite{nawarathna2014abnormal}    &2014 &  $\surd$&  &  & $\surd$& $\surd$ & & & Texture  \\
    Kumar et al. \cite{6051474}    &2012 &  & &  & & & & $\surd$ & Color, Shape and Texture \\
    Ciaccio et al. \cite{ciaccio2013implementation}  &2013 &  & &  & & & $\surd$ &  & Texture \\
    Zhao et al. \cite{zhao2015general} & 2015 &  & $\surd$ &  & & $\surd$ & & & Color, Shape and Texture \\
    
    \hline
    \end{tabular}
}
\caption{Overview of CAD systems.}
\label{tab:lit_review}
\end{table*}

\paragraph{Color Features}
Color is a very useful feature to detect certain kind of pathologies or GI events, such as bleeding, turbid or bubbles. Figuereido et al. in \cite{figueiredo2013computer} use the second component of CIE Lab colour space together with segmentation and enhancement techniques in order to detect bleeding images. Sanju et al. \cite{Sainju2014} present a method to detect bleeding regions by using statistical features such as mean, standard deviation, entropy, skewness and energy, all derived from the RGB histogram that represents the color probability distribution of the images. Yu et al. in \cite{6497444} group pixels through superpixel segmentation and classify them as bleeding using the red ratio in RGB space. Yeh et al. in \cite{yeh2014bleeding} use the coherence color vector in RGB and HSV color spaces to define the bleeding regions. Segui et al. in \cite{sseguiTurbid} learn a color quantization that is suited for WCE videos and use it in a 64 bin color histogram to discriminate the frames with intestinal content. 

\paragraph{Shape Features} Shape is the most popular feature for the detections of specific events such as polyps and wrinkles. Segui et al. in \cite{sseguiWrinkles} define a centrality descriptor based on Hessian matrix and graph structures. The method finds the star-shape pattern presented in wrinkle frames. Both works in \cite{polypsMamonov} and \cite{silivaPolyp2014} first segment the images and then extract geometric information from polyp candidate areas. Kumar et al. in \cite{6051474} use the MPEG-7 edge histogram descriptor for Crohn's characterization.

\paragraph{Texture Features} Texture features are used for the detection of several pathologies such as Crohn's, Celiac, Ulcers, Polyps and Tumors as well as the detection of bubbles. Local Binary Patterns (LBP) are used by Nawarathna et al. in \cite{nawarathna2014abnormal} to detect several abnormalities such as erythema, polyps or ulcers. Eid et al. \cite{Eid2013} introduce the multi-resolution discrete curvelet transform \cite{candes2006fast} to detect ulcer images. Ciaccio et al. \cite{ciaccio2013implementation} use average and standard deviation in gray-scale level of several small sub-windows of the images to characterize Celiac disease.  MPEG-7 texture descriptor is used by Kumar et al. in \cite{6051474} for the characterization of Crohn's disease and by Zhao et al. in \cite{zhao2015general} to characterize intestinal content as well as different lesions as for instance polyps.

\section{Frame-based motility events}

\begin{figure}
        \centering
        
        \begin{subfigure}[b]{0.5\textwidth}
                \includegraphics[width=0.11\textwidth]{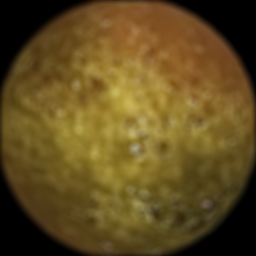}
                \includegraphics[width=0.11\textwidth]{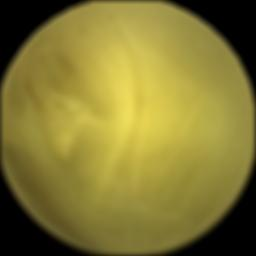}
                \includegraphics[width=0.11\textwidth]{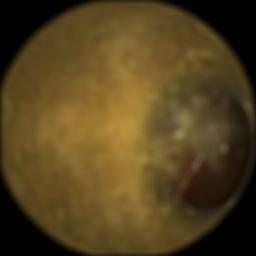}
                \includegraphics[width=0.11\textwidth]{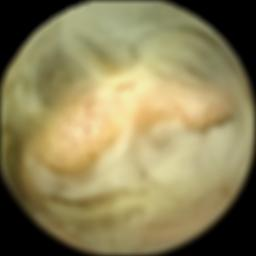}
                \includegraphics[width=0.11\textwidth]{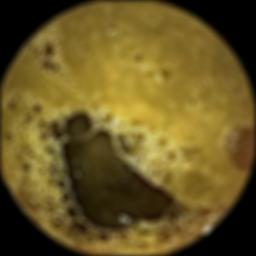}
                \includegraphics[width=0.11\textwidth]{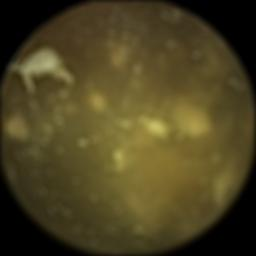}
                \includegraphics[width=0.11\textwidth]{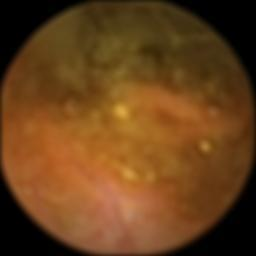}
                \includegraphics[width=0.11\textwidth]{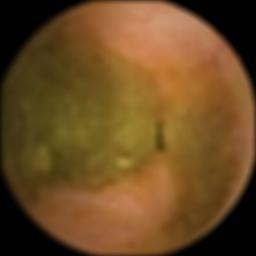}
                \caption{Turbid}\label{fig:dataset_1}
        \end{subfigure}%
        
        \begin{subfigure}[b]{0.5\textwidth}
                \includegraphics[width=0.11\textwidth]{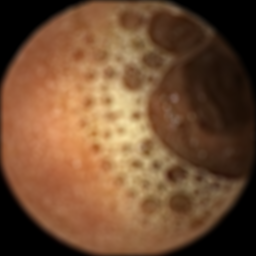}
                \includegraphics[width=0.11\textwidth]{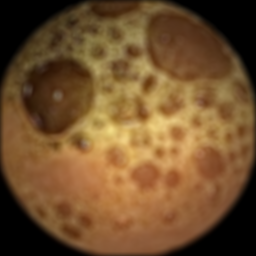}
                \includegraphics[width=0.11\textwidth]{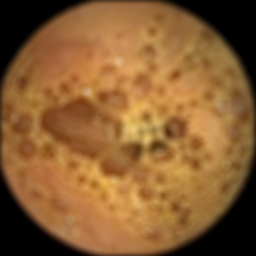}
                \includegraphics[width=0.11\textwidth]{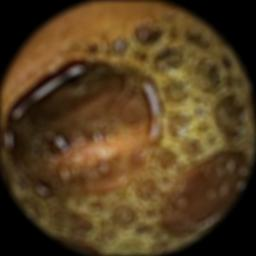}
                \includegraphics[width=0.11\textwidth]{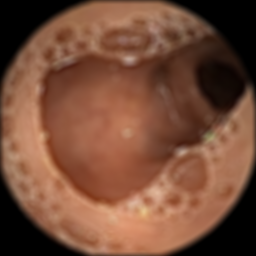}
                \includegraphics[width=0.11\textwidth]{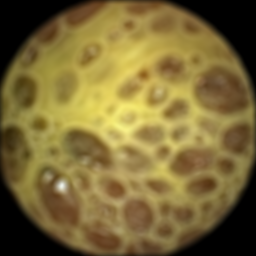}
                \includegraphics[width=0.11\textwidth]{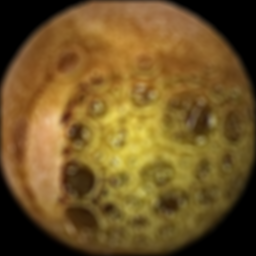}
                \includegraphics[width=0.11\textwidth]{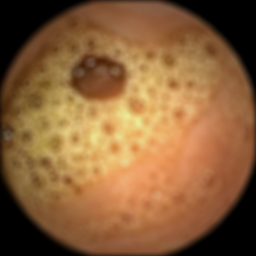}
                \caption{Bubbles}\label{fig:dataset_2}
        \end{subfigure}%
        
        \begin{subfigure}[b]{0.5\textwidth}
                \includegraphics[width=0.11\textwidth]{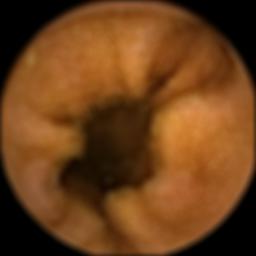}
                \includegraphics[width=0.11\textwidth]{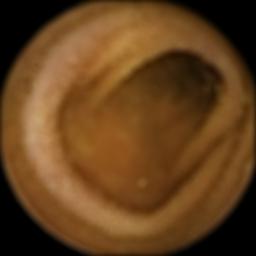}
                \includegraphics[width=0.11\textwidth]{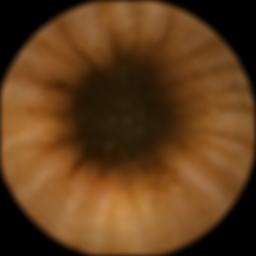}
                \includegraphics[width=0.11\textwidth]{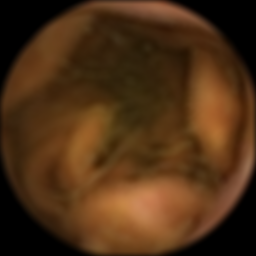}
                \includegraphics[width=0.11\textwidth]{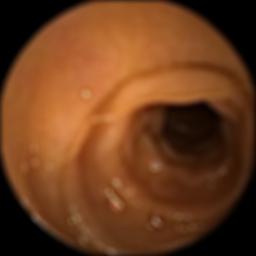}
                \includegraphics[width=0.11\textwidth]{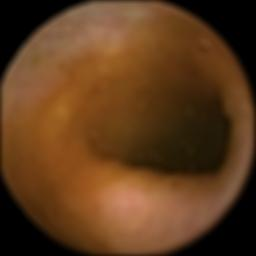}
                \includegraphics[width=0.11\textwidth]{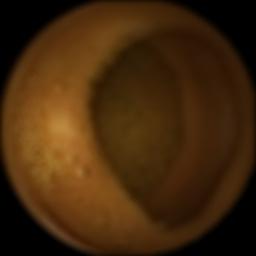}
                \includegraphics[width=0.11\textwidth]{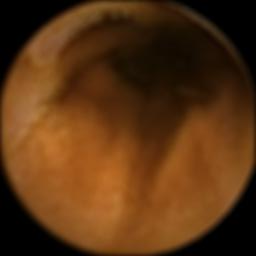}
                \caption{Clear Blob}\label{fig:dataset_3}
        \end{subfigure}%
        
        \begin{subfigure}[b]{0.5\textwidth}
                \includegraphics[width=0.11\textwidth]{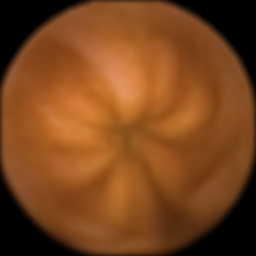}
                \includegraphics[width=0.11\textwidth]{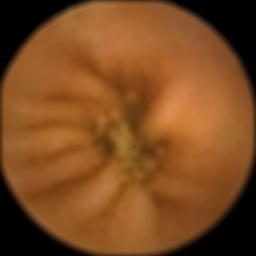}
                \includegraphics[width=0.11\textwidth]{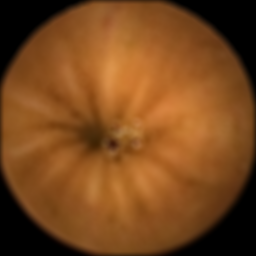}
                \includegraphics[width=0.11\textwidth]{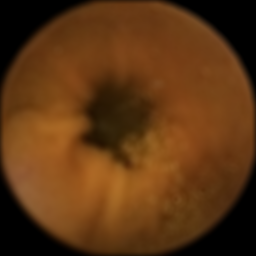}
                \includegraphics[width=0.11\textwidth]{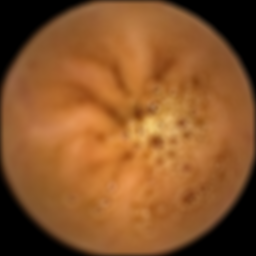}
                \includegraphics[width=0.11\textwidth]{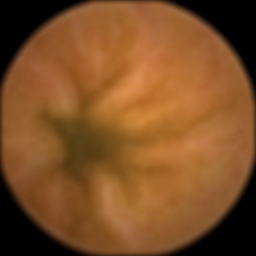}
                \includegraphics[width=0.11\textwidth]{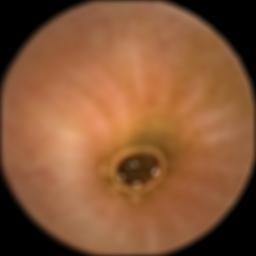}
                \includegraphics[width=0.11\textwidth]{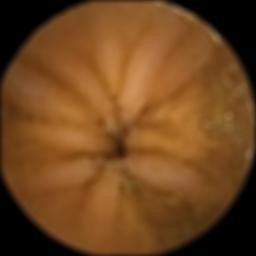}
                \caption{Wrinkles}\label{fig:dataset_4}
        \end{subfigure}%
        
        \begin{subfigure}[b]{0.5\textwidth}
                \includegraphics[width=0.11\textwidth]{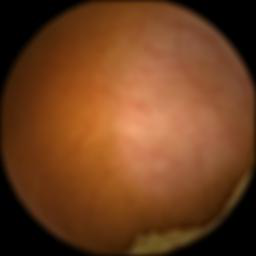}
                \includegraphics[width=0.11\textwidth]{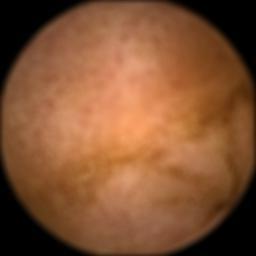}
                \includegraphics[width=0.11\textwidth]{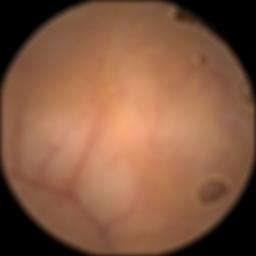}
                \includegraphics[width=0.11\textwidth]{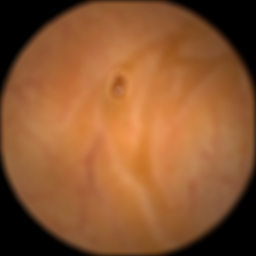}
                \includegraphics[width=0.11\textwidth]{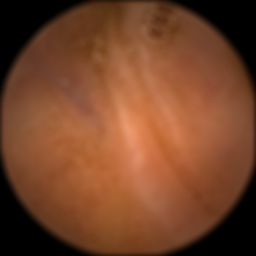}
                \includegraphics[width=0.11\textwidth]{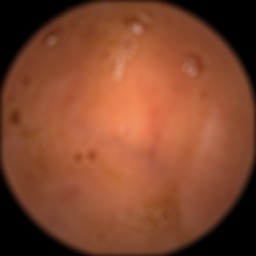}
                \includegraphics[width=0.11\textwidth]{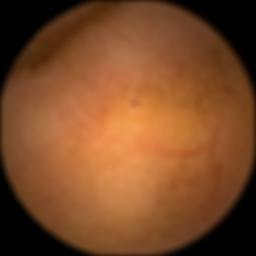}
                \includegraphics[width=0.11\textwidth]{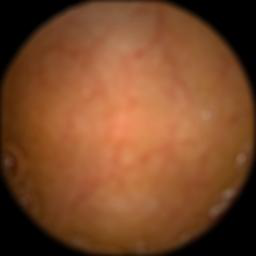}
                \caption{Wall}\label{fig:dataset_5}
        \end{subfigure}%

        \begin{subfigure}[b]{0.5\textwidth}
                \includegraphics[width=0.11\textwidth]{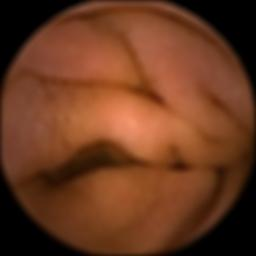}
                \includegraphics[width=0.11\textwidth]{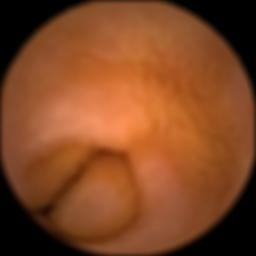}
                \includegraphics[width=0.11\textwidth]{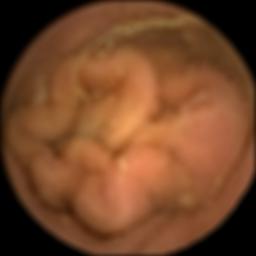}
                \includegraphics[width=0.11\textwidth]{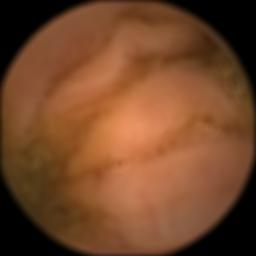}
                \includegraphics[width=0.11\textwidth]{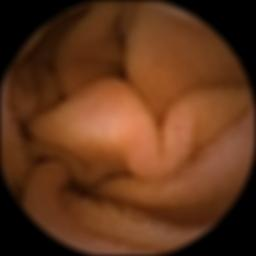}
                \includegraphics[width=0.11\textwidth]{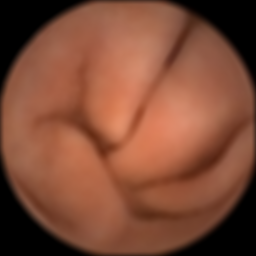}
                \includegraphics[width=0.11\textwidth]{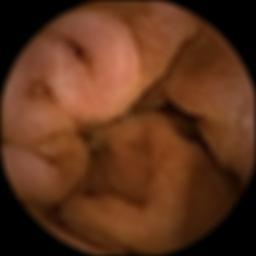}
                \includegraphics[width=0.11\textwidth]{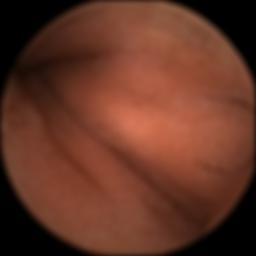}
                \caption{Undefined}\label{fig:dataset_6}
        \end{subfigure}%
        \caption{Eight exemplary images for each of the category in the database.}
        \label{fig:dataset}
\end{figure}

The analysis of visual information in WCE images allows to detect different small intestine diseases: bleeding, Crohn's disease, motility, polyps, etc. Due to the difficulty of getting a large amount of WCE frames with intestinal diseases, we narrowed the problem of WCE frame classification to motility events classification in the small intestine. We defined a set of motility interesting events that can be observed in a single WCE frame \cite{Malagelada2015}:
\begin{itemize}
\item \emph{Turbid} (Figure \ref{fig:dataset_1}): Turbid frames represent food in digestion. These frames usually present a wide range of green-like colors with a homogeneous texture. 
\item \emph{Bubbles} (Figure \ref{fig:dataset_2}): The presence of bubbles is related to agents that reduce surface tension, analogous to a detergent. These frames are usually characterized by the presence of many circular blobs that can be white, yellow or green. 
\item \emph{Clear Blob} (Figure \ref{fig:dataset_3}): These frames display an open intestinal lumen that is usually visible as a variable-size dark blob surrounded by orange-like intestinal wall.
\item \emph{Wrinkles} (Figure \ref{fig:dataset_4}): These frames are characterized by a star-shape pattern produced by the pressure exert by the nerve system and are often present in the central frames of intestinal contractions.
\item \emph{Wall} (Figure \ref{fig:dataset_5}): These images display the orange-like intestinal wall (frames without the presence of the lumen).  
\item \emph{Undefined} (Figure \ref{fig:dataset_6}): These frames correspond to visually undefined clinical events.
\end{itemize}

As a result, our data set of WCE images consists of frames from 50 WCE recordings of small intestine in healthy subjects obtained using the PillCam SB2 (Given Imaging, Ltd., Yoqneam, Israel). During the labeling process, the frames from all 50 videos were randomized and showed to an expert who labeled all of them (until reaching a minimal number of labeled images, namely $20,000$, per each class). 

\begin{figure}[!t]
        \centering
        \begin{subfigure}[b]{0.065\textwidth}
                \includegraphics[width=\textwidth]{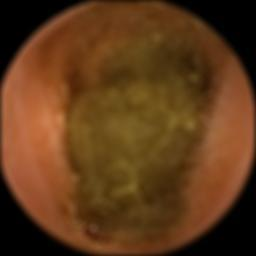}
                \caption{}\label{fig:example_1}
        \end{subfigure}%
        ~ 
        \begin{subfigure}[b]{0.065\textwidth}
                \includegraphics[width=\textwidth]{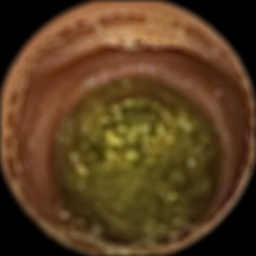}
                \caption{}\label{fig:example_2}
        \end{subfigure}
        ~ 
        \begin{subfigure}[b]{0.065\textwidth}
                \includegraphics[width=\textwidth]{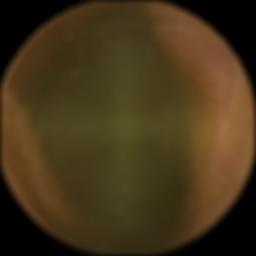}
                \caption{}\label{fig:example_3}
        \end{subfigure}
        ~
        \begin{subfigure}[b]{0.065\textwidth}
                \includegraphics[width=\textwidth]{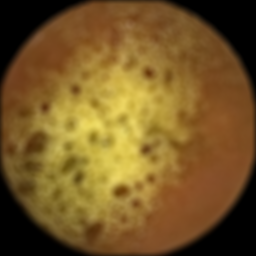}
                \caption{}\label{fig:example_4}
        \end{subfigure}%
        ~ 
        \begin{subfigure}[b]{0.065\textwidth}
                \includegraphics[width=\textwidth]{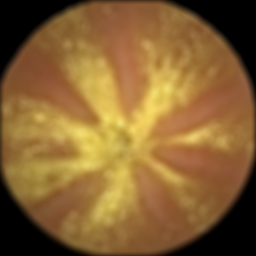}
                \caption{}\label{fig:example_5}
        \end{subfigure}
        ~
        \begin{subfigure}[b]{0.065\textwidth}
                \includegraphics[width=\textwidth]{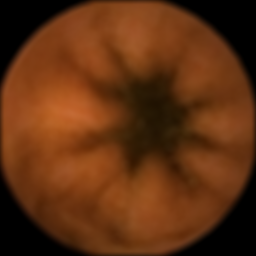}
                \caption{}\label{fig:example_6}
        \end{subfigure}
        
        \caption{Which is the correct label? This figure shows some images that do not have a clear label.}
        \label{fig:difficultImages}
\end{figure}

In order to fully appreciate the complexity and ambiguity of the data set, the reader is invited to review the images displayed in Figure \ref{fig:difficultImages}. For instance, Figures \ref{fig:example_1} and \ref{fig:example_2} are labeled as \emph{clear blob}, although, they could  also be labeled as \emph{turbid}, since the whole lumen is fully covered by turbid. Figure \ref{fig:example_4} is labeled as \emph{turbid}, nonetheless, the bubbles are also present. Figure \ref{fig:example_5} is labeled as \emph{turbid}, yet a clear wrinkle pattern is visible. Finally, Figure \ref{fig:example_6} is labeled as \emph{clear blob}, although it could be considered as \emph{wrinkle}, since the characteristic wrinkle structure is well displayed.

\section{Feature Learning using Convolution Neural Networks}
\label{secMethodology}

Modern Neural Network architectures date from the beginning of 80s \cite{fukushima1980neocognitron}. However, it is not until the very recent years when their use in the computer vision field has fully emerged. Their success has been possible, because of some recent advances related to the training methods of large Convolutional Neural Networks (CNNs) \cite{krizhevsky2012imagenet} as well as to the availability of very large training datasets. CNNs are a variation of the classical Multilayer Perceptron (MLP), which are inspired by biological models. CNN architecture, compared to the MLP, has a much fewer number of connections and parameters by using prior knowledge, such as using weight sharing among local regions of the image.

\subsection{CNN Architecture}
CNNs are defined as a stack of layers with different properties. The lower layers are composed of alternating convolution, normalization and pooling layers, and the upper layers are fully-connected and correspond to traditional neural networks. 

\subsubsection{Convolutional layers}

The input to the first convolutional layer is a $H \times W \times C$ image, where $H$ and $W$ are respectively the height and the width of the image and $C$ is the number of channels, e.g. an RGB image has $C=3$. The input is processed by a set of $k$ convolution filters of size $m \times n \times C$, with $m < H$ and $n < W$ to obtain $k$ features maps of size $H-m+1 \times W-n+1$. At last, an additive bias and a nonlinearity, sigmoidal or ReLU (Recfier Linear Unit), are applied to the feature maps in order to get the final output.

After the first convolutional layer we can stack a number of convolutional layers. Each one of these layers process the feature map produced by the previous layer and produce as output another feature map. 

\subsubsection{Normalization layers}
A normalization layer performs a kind of {\em lateral inhibition} by normalizing over local input regions of the feature maps. Each input value $x$ is divided by $(1+(1/n) \sum_i x_i^2)$, where $n$ is the size of each local region, and the sum is taken over all inputs values located in a region centered at $x$.

\subsubsection{Pooling layers}
After each convolutional or normalization layer there may be a pooling layer. The pooling layer role is to subsample the features maps, pooling over $p \times p$ contiguous regions, to produce a smaller version of each feature map. There are several ways to do this pooling, such as taking the average, the maximum, or a learned linear combination of the elements of the feature map in the block. Our pooling layers will always be max-pooling layers; that is, they take the maximum of the block they are pooling.

\subsubsection{Fully-connected layers}
A fully-connected layer takes all available inputs from the previous layer (be it fully connected, pooling, or convolutional) and connects them to a defined number of outputs through a weight per connection plus a bias. This kind of layer builds a representation where the concept of spatial location makes no sense anymore.

\subsubsection{Classification layer}
The last layer of a deep network is a fully-connected layer with an output that represents the classification problem. In our case, a one-of-many classification task, the chosen output is a probability distribution over classes. To this end, we have used a multinomial logistic loss for training the net, passing real-valued predictions through a softmax to get a probability distribution over classes.

\subsection{Our network architecture}
\label{subsec1}

We have considered one basic architecture plus two variations, all of them 5-layer deep. The variations are introduced in order to test the influence of additional information (priors) on system performance. All these architectures consider the most basic image  features: the 3 channels representing the RGB image. The two variations consider additional information: a channel $L$ representing the Laplacian of the image brightness $I$, and a channel $H$ representing the Hessian of the image brightness $I$. The consideration of priors in a form of $L$ and $H$ is based on the observation that these features showed very good results, when detecting several kinds of WCE frame events \cite{sseguiWrinkles, sseguiTurbid}. 

More formally, $L$ can be easily computed from image derivatives:

$$ L(x,y) = \frac{\partial^2 I}{\partial x^2} +\frac{\partial^2 I}{\partial y^2} $$

$H$ can be derived from the Hessian matrix of every pixel of the image. The Hessian Matrix (HM) is a matrix derived from the second derivatives of the image that summarizes the predominant directions of the local curvatures and their magnitudes in a neighborhood of a pixel:

$$
HM(x,y) = 
\left( \begin{array}{cc}
\frac{\partial^2 I}{\partial x^2} & \frac{\partial^2 I}{\partial x \partial y}  \\
\frac{\partial^2 I}{\partial x^2 \partial y} & \frac{\partial^2 I}{\partial y^2}  \\ \end{array} \right)
$$

Let $\lambda_1$ be the largest eigenvalue by absolute value, $|\lambda_1| > |\lambda_2|$. $|\lambda_1|$ shows the strength of the local image curvature, a concept that can be used to represent foldings of the intestinal wall. To build $H$,  we consider for
every pixel the map represented by $max(0, \lambda_1)$. 

Our basic network input is a $100 \times 100$ pixel RGB image. The lower part of the network is composed of three convolutional layers, along with their corresponding normalization and pooling layers, with $25 \times 25$, $5 \times 5$ and $3 \times 3$ convolution filters respectively.  The dimension of their feature map output is $64$ in all cases. The higher part of the network is composed of two fully connected layer of $512$ neurons. The output layer is a $6$ neuron fully connected layer where each neuron represents a different class.

The first variation of our basic architecture is a three-stream network (as shown in Table \ref{tab:lateStyle}). The input of the streams are: a  $100 \times 100$ pixel RGB image in the first case, a $100 \times 100$ pixel $L$ image for the second case and a $100 \times 100$ pixel $H$ image in the third case. Each stream is composed of three convolutional layers, along with their corresponding normalization and pooling layers. At the end of each stream, the feature maps produced by the last convolutional layers are concatenated. This combination constitutes the input of the fully connected layers.

This architecture can be seen as a {\em late-fusion scheme} for combining color, Hessian and Laplacian features, hence supposing that these features can be considered independent. In this case, the optimal classifier can be built by combining the independently computed feature maps of these three features. 

The second variation we have considered is an early-fusion scheme (see Table \ref{tab:earlyStyle}). In this case, there is only a stream that takes as input a 5-band image composed by concatenating the RGB image, the Hessian and the Laplacian bands. The architecture of this network is the same as the basic network with one exception: the first convolutional layer process a 5-dimensional image instead of a 3-dimensional image.

The number of filters at the convolutional part of the networks have been defined in order to get a comparable representation of the image in terms of parameters. That is, the number of weights that are computed when training all three networks is in the same magnitude order.

\begin{figure*}
\centering
\includegraphics[width=10cm]{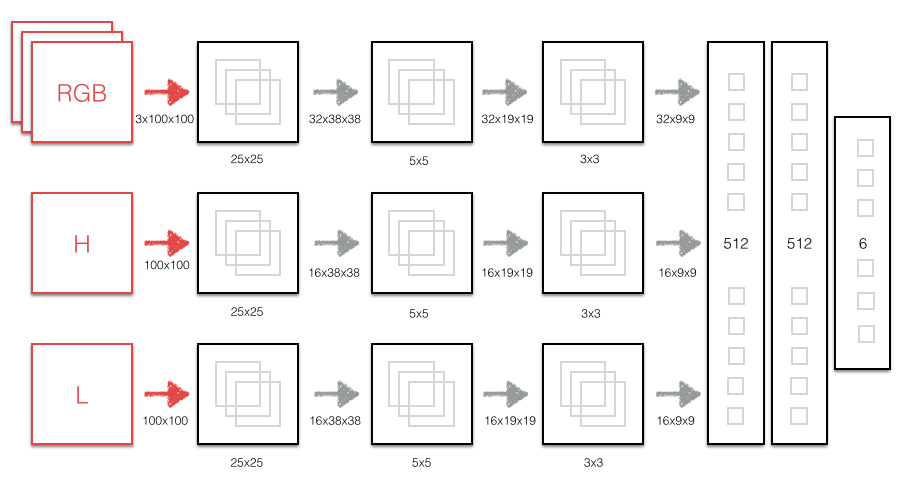}
\caption{The late-fusion architecture is composed of three image streams: The RGB stream, the Hessian stream and the Laplacian stream. Each stream is processed by three convolutional layers with max-pooling and normalization steps. Numbers below arrows show the dimension of the feature maps at every step. Numbers below convolutional layers show the spatial dimensions of their filters. The final steps are three fully connected layers of 512, 512 and 6 neurons, respectively. }
\label{fig:CNN_Architecture}
\end{figure*}

\begin{table}[!ht]
\centering
\scriptsize
\begin{tabular}{  c  c |  c} 

      Layer &    Size      & \# Parameters \\
               \hline
$INPUT_{RGB}$ &    [100x100x3]& \\
$INPUT_H$   &   [100x100x1]& \\
$INPUT_L$   &   [100x100x1]& \\
  		&      \\
RGB CONV25-32& [100x100x32] 		&   (25*25*3)*32 	  =      60,000\\
RGB POOL2-32&  [25x25x32] 	& \\
RGB CONV5-32&  [25x25x32] 		&   (5*5*32)*32 	  =      25,600\\
RGB POOL2-32&  [13x13x32] 	& \\
RGB CONV3-32&  [13x13x32] 		&   (3*3*32)*32 	  =       9,216\\
RGB POOL2-32&  [7x7x32] 		&   \\
&~ & \\
H CONV25-32& [100x100x16] 		&   (25*25*1)*16 	  =      10,000\\
H POOL2-32&  [25x25x16] 	& \\
H CONV5-32&  [25x25x16] 		&   (5*5*16)*16 	  =       6,400\\
H POOL2-32&  [13x13x16] 	& \\
H CONV3-32&  [13x13x16] 		&   (3*3*16)*16 	  =       2,304\\
H POOL2-32&  [7x7x16] 		&   \\
&~ & \\
L CONV25-32& [100x100x16] 		&   (25*25*1)*16 	  =      10,000\\
L POOL2-32&  [25x25x16] 	& \\
L CONV5-32&  [25x25x16] 		&   (5*5*16)*16 	  =       6,400\\
L POOL2-32&  [13x13x16] 	& \\
L CONV3-32&  [13x13x16] 		&   (3*3*16)*16 	  =       2,304\\
L POOL2-32&  [7x7x16] 		&   \\ 
& ~ & \\
CONCAT      & [7x7x64]  & \\
FC&         [1x1x512] 			&   7*7*64*512   	  =    1,605,632\\
FC&         [1x1x512] 			&   512*512 	          =      262,144\\
FC&         [1x1x6]			&   512*6 	  	  =        3,072\\   
\hline

\multicolumn{2}{c}{ Total Number of Parameters:} & \textbf{2M} \\

\end{tabular}

\caption{Late-fusion scheme. Architecture description and the number of parameters.}
\label{tab:lateStyle}
\end{table}

\begin{table}[!ht]
\centering
\scriptsize
\begin{tabular}{  c  c | c} 

         Layer&  Size     & \# Parameters \\
               \hline
$INPUT_{RGBHL}$&     [100x100x5]  		&     \\
CONV25-64& [100x100x64] 		&   (25*25*5)*64 	  =      200,000\\
POOL2-64&  [25x25x64] &	\\
CONV5-64&  [25x25x64] 			&   (5*5*64)*64 	  =      102,400\\
POOL2-64&  [13x13x64] &	\\
CONV3-64&  [13x13x64] 			&   (3*3*64)*64 	  =       36,864\\
POOL2-64&  [7x7x64] 			&   \\
FC&         [1x1x512] 			&   7*7*64*512   	  =    1,605,632\\
FC&         [1x1x512] 			&   512*512 	          =      262,144\\
FC&         [1x1x6]		    	&   512*6 	  	  =        3,072\\   
\hline
\multicolumn{2}{c}{ Total Number of Parameters:} & \textbf{2.2M} \\

\end{tabular}
\caption{Early-fusion scheme. Architecture description and the number of parameters.}
\label{tab:earlyStyle}
\end{table}

\begin{table}[!ht]
\centering
\scriptsize
\begin{tabular}{  c  c | c} 

        Layer   &   Size  & \# Parameters \\
               \hline
$INPUT_{RGBHL}$  & [100x100x5]  		&      \\
CONV3-16 & [100x100x16] 		&   (3*3*5)*16 	  =      720 \\
CONV3-16& [100x100x16]  	&   (3*3*16)*16   =    2,304 \\
POOL2:   &  [50x50x32] 		&    \\
CONV3-32&  [50x50x32] 		&   (3*3*16)*32   =    4,608\\
CONV3-32&  [50x50x32] 		&   (3*3*32)*32   =    9,216\\
POOL2:   &  [25x25x32] 		&   \\
CONV3-64&  [25x25x64] 		&   (3*3*32)*64   =   18,432\\
CONV3-64 & [25x25x64] 		&   (3*3*64)*64   =   36,864\\
POOL2:    & [13x13x64] 		&   \\
CONV3-128 &[13x13x128] 		&   (3*3*64)*128  =   73,728\\
CONV3-128 &[13x13x128] 		&   (3*3*128)*128 =  147,456\\
POOL2:  &    [7x7x128] 		&   \\
FC:      &   [1x1x512] 		&   7*7*128*512   = 3,211,264\\
FC:      &   [1x1x512] 		&   512*512 	  =   262,144\\
FC:      &   [1x1x6]		&   512*6 	  =     3,072\\   
\hline
\multicolumn{2}{c}{ Total Number of Parameters:} & \textbf{3.7M} \\

\end{tabular}

\caption{VGG-style scheme. Architecture description and the number of parameters.}

\label{tab:vggStyle}
\end{table}

Finally, we experiment with VGG-stile architecture \cite{DBLP:journals/corr/SimonyanZ14a}. We build a network with smaller convolutional kernels and with more layers. The architecture is shown in Table \ref{tab:vggStyle}. 

\subsection{CNN Training}
Original PillCam SB2 images have the resolution of $256 \times 256$ pixels. First, we resize all images to $128 \times 128$ pixels then we crop a central $100 \times 100$ part of the image and, finally, we pre-calculate Laplacians and Hessians. The CNN is trained with the open source Convolutional Architecture for Fast Feature Embedding  (CAFFE) environment, presented in \cite{Jia13caffe}. The parameters are initialized with Gaussians (std=1). In order to optimize the network, we use Stochastic Gradient Descent policy with batch size of 128. We start with the learning rate of 0.1 and decrease it every 100k iterations by a factor of 10. The algorithm is stopped after 400k iterations. The system is installed on an Intel® Xeon® Processor E5-2603 with 64GB RAM and a Nvidia Tesla K40 GPU. The network training, for every of the proposed network variation, takes approximately 1 day.

\section{Experimental Results}
\label{secResults}

In this section, we present the experimental results of the proposed system. First, the database is split into two different sets: training and test set with 100k and 20k of the samples (randomly sampled in a stratified way) from the full database, correspondingly. Second, we evaluate the system quantitatively comparing its performance to the state of the art image descriptors. Finally, in order to provide additional understanding of the proposed system, a qualitative analysis is performed.

\subsection{Quantitative results}

Classical image representations (such as GIST, SIFT or COLOR) followed by a Linear Support Vector Machine classifier (available in the sklearn toolbox \cite{scikit-learn}) are used to establish a baseline in our problem. In particular, we use the following image representations:
\begin{itemize}
\item GIST \cite{oliva2001modeling}:  This is a low dimensional representation of the image based on a set of perceptual dimensions (naturalness, openness, roughness, expansion, ruggedness) that
represent the dominant spatial structure of a scene. These dimensions may be reliably estimated using spectral and coarsely localized information.
\item COLOR: Each image is described with 128 color words that are learnt with $k$-means clustering. The training set is a large random sample of $(r,g,b)$ values from WCE videos. 
\item SIFT \cite{Lowe_sift}: Each image is described with 128 SIFT words that are learnt with $k$-means clustering. The training set is a large random sample of image frames from WCE videos.
\item SIFT+COLOR: A concatenation of SIFT and COLOR descriptor is used.
\item GIST+SIFT+COLOR: A concatenation of all above features.
\end{itemize}

In the experiments, we use the following notations for different CNN modalities (for details, see Section IV): $CNN_{RGB}$ refers to our basic network, with a RGB image as input; $CNN^l_{RGBHL}$ refers to the late-fusion network, $CNN^e_{RGBHL}$ to the early-fusion network and $CNN^{vgg}_{RGBHL}$ is a VGG-stile network. 

The first experiment is designed to compare the classification results we can obtain with different image features: GIST, SIFT , COLOR and CNN.  As it can be seen in Table \ref{tab:comparisionResults}, the best results are obtained with the $CNN_{RGB}$ system with a mean accuracy of 96\%. It is worth noticing that $CNN_{RGB}$ outperforms classical image representations in all categories, with the best performance for \emph{wall} class (99.0\%) and the worst performance for \emph{turbid} class (92.2\%). Then, when comparing SIFT, GIST and COLOR, we can observe that GIST descriptor achieves the best results with a mean accuracy of 78.0\%. Not surprisingly, GIST descriptor performs well for \emph{wall}, \emph{wrinkles} and \emph{bubbles}, while COLOR descriptor achieves good results only for the \emph{wall} class. The second best result for our database is obtained by concatenating all classical image representations (GIST with SIFT and COLOR).

\begin{table*}[!ht]
\centering
\resizebox{13cm}{!} {
\begin{tabular}{l*{10}{c}r} 
               & GIST & SIFT & COLOR & SIFT+COLOR & GIST+SIFT+COLOR & LinearSVM (CNN feat) & $CNN_{RGB}$ \\
\hline
  Wall        & 89.9    & 49.15 & 92.2 &   95.2   &  90.7   & 94.31   & 99.0 \\
  Wrinkles    & 88.2    & 67.0  & 36.9 &   75.7   &  82.2   & 90.86   & 95.9 \\
  Bubbles     & 95.9    & 39.75 & 65.4 &   82.5   &  91.2   & 95.65   & 97.1 \\
  Turbid      & 65.1    & 41.1  & 50.2 &   58.6   &  80.2   & 84.88   & 92.2 \\
  Clear Blob  & 70.1    & 77.3  & 80.9 &   77.7   &  77.4   & 91.16   & 97.7 \\
  Undefined   & 58.4    & 21.25 & 61.0 &   39.1   &  74.7   & 84.73   & 92.9 \\
  \hline
  \hline
  Mean       & \textbf{78.0}    & \textbf{50.3}  & \textbf{64.4} & \textbf{71.5} & \textbf{82.8} & \textbf{90.2} & \textbf{96.01} \\
\end{tabular}
}
\caption{Comparison of accuracy obtained for frame classification in WCE data. The numbers represent percentages.}
\label{tab:comparisionResults}
\end{table*}

In order to understand better the results for the most difficult class, the \emph{turbid} class, the confusion matrix for $CNN_{RGB}$ is presented in Table \ref{tab:resCNN}. Not surprisingly, the turbid class is miss-classified from time to time as \emph{bubble} class (5.9\%), but even an expert can have difficulty in distinguishing turbid from bubbles in some images. Another interesting case is the \emph{undefined} class, which is miss-classified for \emph{clear blob} 3\% of the time and for \emph{wall} 1\% of the time.

\begin{table}[!ht]
\centering
\resizebox{9cm}{!} {
\noindent\begin{tabular}{c | c | c | c | c | c | c|}
 \multicolumn{1}{c}{} & \multicolumn{1}{c}{Wall} & \multicolumn{1}{c}{Wrinkles} 
  & \multicolumn{1}{c}{Bubbles} & \multicolumn{1}{c}{Turbid} & \multicolumn{1}{c}{Clear Blob} 
  & \multicolumn{1}{c}{Undefined} \\ \hline
 Wall       & 99.0 & 0.0 & 0.0 & 0.0 & 0.0 & 0.8 \\  \hline
 Wrinkles   & 0.7 & 95.9 & 0.2 & 0.5 & 1.2 & 1.6 \\ \hline
 Bubbles    & 0.1 & 0.0 & 97.1 & 2.6 & 0.0 & 0.1 \\ \hline
 Turbid     & 0.3 & 0.1 & 5.9 & 92.2 & 1.0 & 0.3 \\ \hline
 Clear Blob & 0.0 & 0.3 & 0.0 & 0.5 & 97.7 & 1.2 \\ \hline
 Undefined  & 2.1 & 1.3 & 0.1 & 0.4 & 3.0 & 92.9 \\ \hline
\end{tabular}
\par\bigskip
}
\caption{Confusion matrix for our method $CNN_{RGB}$. The numbers represent percentages.}
\label{tab:resCNN}
\end{table}

In the second experiment, we evaluate the impact the number of images in the training set has on the system performance. In order to do so, we randomly subsample each class in the training set. As a result, we obtain three training sets of 1k, 10k and 100k examples. 

Moreover, we perform experiments to see if providing additional information to our system improves its performance. In particular, we measure the accuracy gain of the system by adding Laplacian and Hessian information to the network input (as described in Section IV). The results of the experiments are displayed in Table \ref{tab:fusion}. As it can be seen, the $CNN^l_{RGBHL}$ is the winner architecture when a relatively small amount of images is used for training (1k).

Increasing the number of samples in the training set has a positive effect on the system performance for all architectures, which reaches the accuracy of 93\% for 10k and 96\% for 100k images. Finally, we can observe that adding additional information as an input to the network has a positive impact on the system performance only for small training sets, where we observe an accuracy improvement of 2.6\% for $CNN^l_{RGBHL}$ (w.r.t. $CNN_{RGB}$). This is not surprising, since using additional information can be seen as a prior, which is especially important, when the data set is small. As we increase the size of the training set, the prior has lower impact on the final system performance. From the results, we obtained, it is difficult to determine which merging scheme is better. However, it looks like the late-fusion has slightly better performance (0.4\% accuracy increase w.r.t early-fusion for 1k training set). Finally, we evaluate VGG-like architecture on WCE data $CNN^{vgg}_{RGBHL}$. We observe that $CNN^{vgg}_{RGBHL}$ has similar results to $CNN^l_{RGBHL}$ and to $CNN^e_{RGBHL}$.

\begin{table}[!ht]
\centering
\resizebox{6cm}{!} {
\noindent\begin{tabular}{c | c | c | c | c | c | c |}
 \multicolumn{1}{c}{N=} & \multicolumn{1}{c}{1k} & \multicolumn{1}{c}{10k} 
  & \multicolumn{1}{c}{100k} \\ \hline
 $CNN_{RGB} $      & 83.78 & 92.93 & 96.01  \\  \hline
 $CNN^e_{RGBHL} $  & 85.97 & 93.19 & 96.12  \\ \hline
 $CNN^l_{RGBHL}$   & 86.35 & 92.90 & 96.19  \\ \hline
 $CNN^{vgg}_{RGBHL} $& 86.62 & 92.80 & 96.22  \\ \hline
\end{tabular}
\par\bigskip
}
\caption{Introducing additional information. The numbers represent classification accuracy in percentages.}
\label{tab:fusion}
\end{table}

\subsection{Qualitative results}

In this section, a qualitative analysis of the results is performed. First, the filters learned by $CNN_{RGB}$ and $CNN^l_{RGBHL}$ networks are displayed. Second, a visual evaluation of the learned representation is done. Finally, we show where our system fails. 

\begin{figure*}[!ht]
    \centering
    \begin{subfigure}[b]{0.22\textwidth}
        \includegraphics[width=\textwidth]{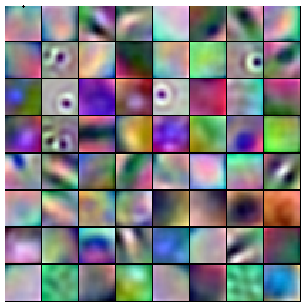}
        \caption{}
        \label{fig:1_layers_conv_RGB}
    \end{subfigure}
    \begin{subfigure}[b]{0.22\textwidth}
        \includegraphics[width=\textwidth]{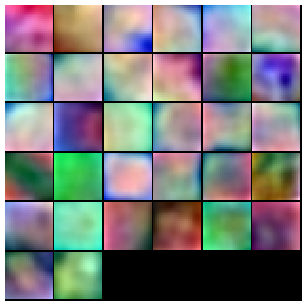}
        \caption{}
        \label{fig:1_layers_conv_RGBHL1}
    \end{subfigure}
     \begin{subfigure}[b]{0.22\textwidth}
        \includegraphics[width=\textwidth]{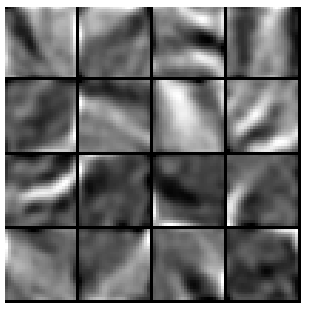}
        \caption{}
        \label{fig:1_layers_conv_RGBHL2}
    \end{subfigure}
     \begin{subfigure}[b]{0.22\textwidth}
        \includegraphics[width=\textwidth]{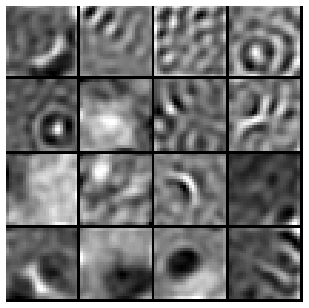}
        \caption{}
        \label{fig:1_layers_conv_RGBHL3}
    \end{subfigure}
    \caption{Visualization of the learnt filters: (a) 64 filters from $CNN_{RGB}$, (b) 32 color filters from $CNN^l_{RGBHL}$, (c) 16 Laplacian filters from $CNN^l_{RGBHL}$, (d) 16 Hessian filters from $CNN^l_{RGBHL}$. All filters are $25 \times 25$ pixels. The figure is best seen in color.}
    \label{fig:1_layers_conv}
\end{figure*}

In order to understand better the differences between $CNN_{RGB}$ and $CNN^l_{RGBHL}$ architectures, we analyse the filters form the network's first layer. The learnt filters (for the training set of 100k examples) are displayed in Figure \ref{fig:1_layers_conv}. Figure \ref{fig:1_layers_conv_RGB} shows filters form $CNN_{RGB}$, while Figures \ref{fig:1_layers_conv_RGBHL1}, \ref{fig:1_layers_conv_RGBHL2} and \ref{fig:1_layers_conv_RGBHL3} show the resulting kernels from $CNN^l_{RGBHL}$. Note that in both cases, the number of parameters distributed along the filters is constant. As it can be seen, filters from Figure \ref{fig:1_layers_conv_RGB} combine both color information and texture. However, if we add additional streams to the network, the system uses RGB channels to learn color information, and Laplacian and Hessian streams to learn the structure and the texture present in the WCE images. 

\begin{figure}[!ht]
\centering
\includegraphics[width=7.5cm]{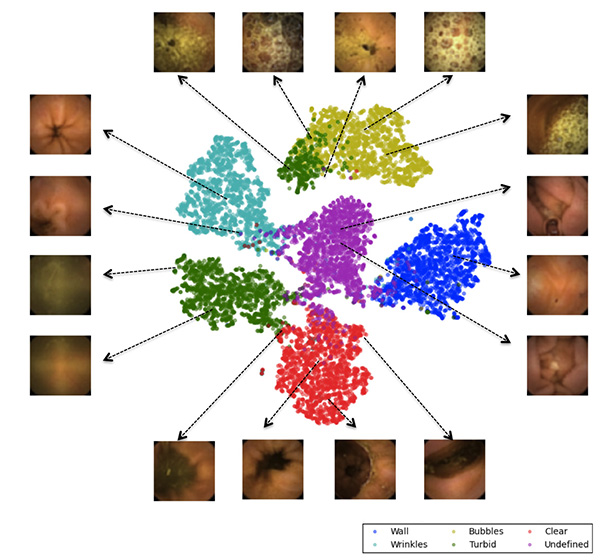}
\caption{Visualization of the class distribution using t-SNE, the corresponding images are shown for some points. Figure is best seen in color.}
\label{fig:class}
\end{figure}

In the next experiment, we visualize the last fully connected layer for all images in the training set for $CNN_{RGB}$ network learnt with 100k images. In order to project the 512 dimensional vector to 2 dimensions, we use the t-SNE algorithm \cite{t_SNE}. The results are presented in Figure \ref{fig:class}. The colors of the points indicate the class assigned by an expert. Two interesting observations can be made: 1) there are two types of \emph{turbid} frames (marked with green color in the figure), \emph{turbid} can either be smooth with small amount of texture or be mixed with bubble like structures (e.g. images in the first row of Figure \ref{fig:class}); 2) the images that are at intersection of two classes tend to present characteristics of both classes (borderline cases, e.g. see left bottom image in Figure \ref{fig:class} that present \emph{clear blob} in the foreground with \emph{turbid} content in the background of the lumen).

\begin{figure*}[!ht]
\centering
\includegraphics[width=12cm]{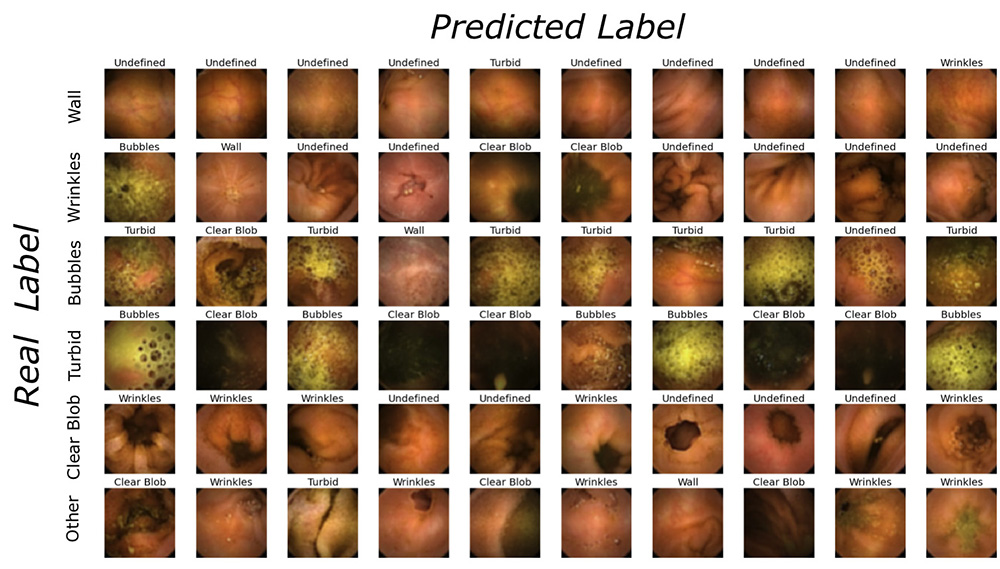}
\caption{Images, where the proposed system fails. Each row contains sample images from one event and at the top of the images, the predicted label is printed. }
\label{fig:negatives}
\end{figure*}

Finally, some failures of the system (100k training set, $CNN_{RGB}$) are presented in Figure \ref{fig:negatives}. Each row shows 10 images from one class that are miss-classified by the system. This Figure \ref{fig:negatives} complements Figure \ref{fig:class} and provides additional information about the system's errors. For example, in many cases the frontier between intestinal content and bubbles is not clear (see third and fourth row of Figure \ref{fig:negatives}), or, while looking at fifth row, it can be seen that some clear blobs have wrinkle-like structure.

\section{Conclusions}

In this paper, we have presented a generic feature descriptor for the classification of WCE images that achieves very good results 96\% of accuracy beating the second best method by almost 14\%. In order to build our system, several steps needed to be done. First, a large data set of labeled images was built; second, we designed a CNN-like architecture using as an input only color images or color images with some additional information; finally, we performed an exhaustive validation of the proposed method. 

From the results presented in Section V, several interesting observations can be made regarding the nature of CNN-like models. First, that generic feature learning works remarkably well for WCE data; even for relatively small number of training images it outperforms classical image representation techniques. Moreover, the more data we have the better results can be obtained. In our problem, with 100k training images we reached outstanding accuracy of 96\%. Second, we investigated the impact of priors on the system accuracy. In our system, we added priors in a form of Hessian or Laplacian image description and we observed that this additional information is especially helpful with small amount of data, for the training set with 1k examples an accuracy improvement of 2.6\% was recorded. While increasing the size of the training set, the impact of priors decreases. Therefore, one way of applying CNN-like models to small data sets (like in medical imaging community) is to design proper priors which could be added directly to a CNN-like model. Finally, it is worth remarking that creating a good quality labeling of large amount of medical data is still a very laborious and expensive task.

As for future work, it might be interesting to build a generic system (not limited to intestinal motility analysis) that would encapsulate all currently used WCE clinical applications. In order to do so, large collections of frames for each application should be collected and labeled. Moreover, we would like to evaluate how well the proposed method behaves when it is integrated with a system for abnormal motility characterization.

\section*{Acknowledgements}
This work was supported in part by a research grant from
Given Imaging Ltd., Yoqneam Israel, as well as by Spanish
MINECO/EU Grant TIN2012-38187-C03 and SGR 1219. MD has received funding from the People Programme of the EU’s 7th Framework Programme under REA grant agreement no. 607652 (ITN NeuroGut). We gratefully acknowledge the support of NVIDIA Corporation with the donation of a Tesla K40 GPU used for this research.

\bibliographystyle{elsarticle-num}
\bibliography{sample}

\end{document}